Robert Nowotniak*, Jacek Kucharski**

# Building Blocks Propagation in Quantum-Inspired Genetic Algorithm

## 1. Introduction

This paper presents an analysis of building blocks propagation in Quantum-Inspired Genetic Algorithm[8],which belongs to a new class of metaheuristics drawing their inspiration from both biological evolution and unitary evolution of quantum systems. The expected number of quantum chromosomes matching a schema has been analyzed and a random variable corresponding to this issue has been introduced. The results have been compared with Simple Genetic Algorithm. Also, it has been presented how selected binary quantum chromosomes cover a domain of one-dimensional fitness function.

The *quantum-inspired genetic algorithms* are located in the intersection of two subfields of computer science: quantum computing and evolutionary computing. Quantum computing is a new branch of computer science concerning applications of unique quantum mechanical effects to solving computational problems[16]. By application of such unique quantum mechanical phenomena it is possible to solve selected computational problems extremely efficiently. Contrary to "true" quantum algorithms (e.g. Grover's search algorithm[5] or Shor's factorization algorithm[17]), the considered algorithms do not require a functional quantum computer for their efficient implementation. Instead, the algorithms exploit additional level of randomness inspired by concepts and principles drawn from quantum mechanical systems, such as qubits or superposition of states. The first proposal of evolutionary algorithm based on the concepts and principles of quantum computing was presented in [15] and this area is still intensively studied nowadays. To simplify the notation, the algorithms will be denoted henceforth in this paper as *quantum genetic algorithms*. No real quantum-level hardware is required for their efficient implementation, however, a possibility of potential implementation of the algorithms on a quantum computer is quite probable in the future.

Recent years have witness successful applications of quantum-inspired evolutionary algorithms in variety of different areas, including image processing[11,18], flow shop scheduling problems[6,7,19], network design problems[20,21], thermal unit commitment[10], power system optimization[1] and many others. Consequently, it has been dem-

* MSc, Computer Engineering Department, Technical University of Łódź

** MSc, PhD, DSc, Computer Engineering Department, Technical University of Łódź

onstrated that quantum-inspired genetic algorithms are capable to outperform classical metaheuristics for a wide range of problems.

Research in the area of quantum genetic algorithms has been started in late 1990s and the algorithms are intensively studied nowadays. However, the theoretical aspect of the algorithms has not been studied with due attention. Strong theoretical foundations of an evolutionary algorithm play a substantial role in its possible applications. The theoretical analysis allows explaining why and how the evolutionary algorithm works. Also, it can be helpful in tuning parameters of the algorithm and identifying the problems in which the algorithm performs best. It makes research on theoretical aspects of quantum evolutionary algorithms highly justified. One possible approach to theoretical analysis of classical genetic algorithms is the Holland's schema theorem[9]. According to this theory, in Simple Genetic Algorithm short, low order, above average schemata receive exponentially increasing trials in subsequent generations and below average schemata receive exponentially decreasing trials. However, it has not been verified yet if the advantage of the quantum genetic algorithms over classical genetic algorithms has its origins in better building blocks propagation. These deficiencies in the significant, theoretical aspect of the quantum-inspired genetic algorithms were the main motivation of this article. The aim of this article was to improve the theoretical foundations of the algorithms by creating a new method of analysis. The proposed method allows comparing how building blocks propagate in quantum-inspired genetic algorithms in comparison to classical genetic algorithms.

This paper is organized as follows. In section 2 the Quantum-Inspired Genetic Algorithm has been presented. In section 3 schemata for binary quantum chromosomes have been considered and a random variable has been introduced, corresponding to the number of binary quantum chromosomes matching the given schema in a quantum population. It is the main contribution of this paper. In section 4 conditions of conducted numerical experiment have been described. In the experiment a comparison of Simple Genetic Algorithm and Quantum-Inspired Genetic Algorithm for selected test problem has been presented. In section 5 an analysis of building block propagation in the test case has been performed and the results have been presented. In section 6 the results have been discussed and it has been presented how selected binary quantum chromosomes "cover" a domain of one-dimensional fitness function. In section 7 the article has been briefly summarized and conclusions have been drawn.

## 2. Quantum-Inspired Genetic Algorithm

The Quantum-Inspired Genetic Algorithms (QIGA) has been proposed in [8] and the algorithm has been briefly presented in this section. In the algorithm a novel representation of solutions is employed. Genes are modelled upon the concept of qubits, which brings an additional element of randomness and a "new dimension" into the algorithm. The *qubit* is a basic unit of quantum information. It is a normalised vector in a two dimensional vector space spanned by the base vectors $|0\rangle$ and $|1\rangle$, as given in equation (1):

$$|\Psi\rangle = \alpha|0\rangle + \beta|1\rangle \tag{1}$$

where: $\alpha, \beta \in \mathbf{C}$, $|0\rangle = [1 \quad 0]^T$, $|1\rangle = [0 \quad 1]^T$ and $|\alpha|^2 + |\beta|^2 = 1$.

Observation of the qubit $|\Psi\rangle$ yields a value 0 with probability $|\alpha|^2$ and value 1 with probability $|\beta|^2$. With some simplifications (imaginary part omitted) a state of qubit can be illustrated by a normalized vector on the two-dimensional space as in figure 1. Entire solutions are represented as *binary quantum chromosomes*, encoded as:

$$q = \begin{bmatrix} \alpha_1 & \alpha_2 & \cdots & \alpha_m \\ \beta_1 & \beta_2 & \cdots & \beta_m \end{bmatrix} \tag{2}$$

where each column corresponds to binary quantum gene $|\Psi\rangle_1, ..., |\Psi\rangle_m$. Hence, a state of quantum population $Q=\{q_1, q_2, ..., q_N\}$ can be simply illustrated by a matrix of vectors, which has been presented in figure 2. Each row in figure 2 corresponds to binary quantum chromosome, as in (2).

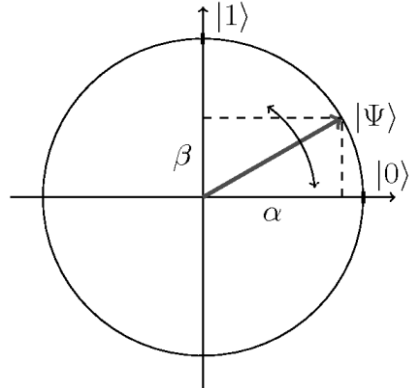

**Fig. 1.** Geometric illustration of binary quantum gene, represented as qubit with its imaginary part omitted. Along with increase of the angle between the vector $|\Psi\rangle$ and the horizontal axis, the probability of observing value 1 grows.

During a phenotype creation states of all genes in quantum chromosomes are *observed*, i.e. the search space is sampled with respect to the probability distribution encoded in the quantum chromosomes. The mapping between binary quantum gene $|\Psi\rangle$ and probability distribution is given as: $\Pr_{|\Psi\rangle}(\{0\}) = |\alpha|^2$, $\Pr_{|\Psi\rangle}(\{1\}) = |\beta|^2$. Obviously, $\Pr_{|\Psi\rangle}(\{0,1\}) = 1$.

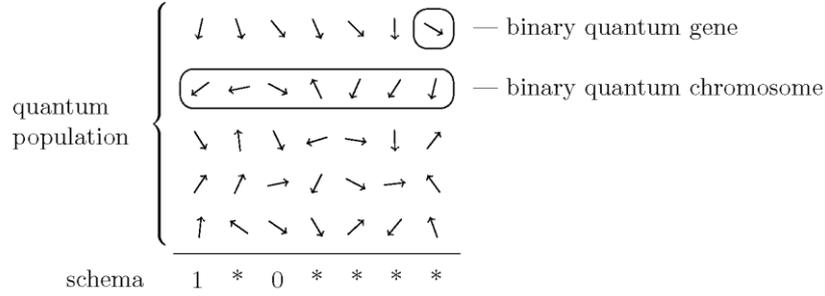

**Fig. 2.** Illustration of binary quantum population. Each arrow represent a state of a quantum gene.

The full pseudocode of Quantum-Inspired Genetic Algorithms[8] is given in figure 3. In the beginning of the algorithm the genes of all individuals in the quantum population $Q(0)$ are initialized with linear superposition $\left(\frac{\sqrt{2}}{2}|0\rangle + \frac{\sqrt{2}}{2}|1\rangle\right)$. This is equivalent to uniform distribution of sampling the search space. The evaluation of individual's fitness in the algorithm is performed by observation of $Q$ yielding $P$. In this step the search space is sampled with respect to the probability distributions encoded by quantum individuals $Q(t)$. The genetic operators applied in the algorithm are based on *quantum rotation gates,* which rotate state vectors in the quantum gene state space. The quantum gate $U(\theta)$ rotates the state of quantum gene $|\Psi\rangle$ by the angle $\theta$. Thus, states of quantum genes are updated according to the following principle of quantum computing:

$$|\Psi\rangle_{t+1} = U(\theta)|\Psi\rangle_t = \begin{bmatrix} \cos(\theta) & -\sin(\theta) \\ \sin(\theta) & \cos(\theta) \end{bmatrix} \begin{bmatrix} \alpha \\ \beta \end{bmatrix} \quad (3)$$

where:

$\theta$ – angle of rotation in the qubit state space

$t$ – iteration number

In recent years several extensions of the algorithm have been proposed. The proposal of other quantum genetic algorithms include GAQPR[2], vQEA[3] and NQEA[13].

```
procedure QIGA
begin
    t←0
    initialize Q(0)
    make P(0) by observing Q(0)
    evaluate P(0)
    store the best solution among P(0)
    while not termination-criterion do
        t←t+1
        make P(t) by observing Q(t-1) population
        evaluate P(t)
        update Q(t) using quantum gates U(θ_t)
        store the best solution among P(t)
    end while
end
```

**Fig. 3.** Pseudocode of Quantum-Inspired Genetic Algorithm

## 3. Schemata and Binary Quantum Chromosomes

In this section a concept of schema for binary quantum chromosomes has been discussed. Let $S$ denote a schema of length $m$, i.e. a word over the ternary alphabet $\{0, 1, *\}$, where the wildcard $*$ denotes an arbitrary element of the alleles set $G = \{0, 1\}$. Also, let $\Pr_g = F_G \mapsto [0,1]$ denote a probability distribution encoded in the binary quantum gene $|g\rangle = \alpha|0\rangle + \beta|1\rangle$. Let $q = g_1g_2...g_m$ denote a binary quantum chromosome of $m$ genes.

For the beginning of our reasoning let us consider the situation when the binary quantum chromosome $g_1g_2g_3$ matches the schema $S = 10*$. It happens when value 1 is observed in $g_1$, value 0 is observed in $g_2$ and observation of $g_3$ can yield an arbitrary value, respectively. Because of random nature of genes in QIGA, a quantum chromosome matches a schema with certain probability. Since quantum genes in quantum-inspired genetic algorithms are independent, i.e. they are modelling two-level isolated quantum systems, the probability that the binary quantum chromosome $q$ matches the schema $S$ is a product of probabilities that observations of consequent genes yielded values corresponding to the elements of the $S$ schema. In other words, the probability $M(q,S)$ that the binary quantum chromosome $q$ matches $S$ is given by the formula[1]:

---

[1] Also, assume that $* = \{0,1\}$. Then: $\Pr_g(*) = \Pr(\{0,1\}) = 1$

$$M(q, S) = \prod_{i=1}^{m} \Pr_{gi}(\{S[i]\}) \tag{4}$$

where:

- $S$ – schema of $m$ elements
- $q$ – binary quantum chromosome of $m$ elements
- $S[i]$ – element of the $S$ schema at the position $i$.

For example, let us consider the binary quantum chromosome $q = g_1g_2g_3 = 10\,(\frac{1}{2}|0\rangle + \frac{\sqrt{3}}{2}|1\rangle)$ and its match to the five schemata: ***, *0*, 10*, **0 and 1*1. The schemata and probabilities of matches have been presented in table 1. The chromosome $q$ matches the schema $S_1 = ***$ regardless of the values observed in quantum genes $g_1g_2g_3$, therefore the probability $M(q,S_1)$ equals 1, according with (4). Also, it is unconditional that $10\,(\frac{1}{2}|0\rangle + \frac{\sqrt{3}}{2}|1\rangle)$ matches *0* and 10*, therefore $M(q,S_2) = 1$ and $M(q,S_3) = 1$. The chromosome $q$ matches the schema **0, if and only if the value 0 are observed in quantum gene $g_3$. The probability of such event is given by $\left|\frac{1}{2}\right|^2$, therefore $M(q,S_4) = \frac{1}{4}$. Finally, the quantum chromosome matches the schema 1*1, if value 1 is observed in genes $g_1$ and $g_3$. The probability of this event is $M(q,S_5) = 1 \cdot \left|\frac{\sqrt{3}}{2}\right|^2 = \frac{3}{4}$.

**Table 1**

|  | **1** | **0** | $\frac{1}{2}|0\rangle + \frac{\sqrt{3}}{2}|1\rangle$ | **M(q,S)** |
|---|---|---|---|---|
| $S_1$ | * | * | * | 1 |
| $S_2$ | * | 0 | * | 1 |
| $S_3$ | 1 | 0 | * | 1 |
| $S_4$ | * | * | 0 | $\frac{1}{4}$ |
| $S_5$ | 1 | * | 1 | $\frac{3}{4}$ |

The matches of binary quantum chromosome $q = g_1g_2g_3 = 10\,(\frac{1}{2}|0\rangle + \frac{\sqrt{3}}{2}|1\rangle)$ for five different schemata. The probability M(q, $S_1$) that $q$ matches $S_2 = *0*$ equals 1. The probability M(q, $S_5$) that $q$ matches $S_5 = 1*1$ equals $\frac{3}{4}$ etc.

Next, let us introduce a random variable $L$ corresponding to the number of quantum chromosomes matching the schema $S$ in a quantum population $Q = \{q_1, q_2, ..., q_N\}$. To shorten the notation, let us denote the probabilities that the schema $S$ has been or has not been matched, respectively, in an observation of quantum chromosome $q_i$ as $D_i^+ = M(q_i, S)$ and $D_i^- = (1 - M(q_i, S))$.

Now, let us calculate the expected number of binary quantum chromosomes matching the schema $S$ in the quantum population $Q$. Based on the definition of the expected value of a random variable:

$$E(L) = 0 \cdot \left[ D_1^- D_2^- \cdot \ldots \cdot D_N^- \right]$$

$$+ 1 \cdot \underbrace{\left[ D_1^+ D_2^- \cdot \ldots \cdot D_N^- + D_1^- D_2^+ \cdot \ldots \cdot D_N^- + \ldots + D_1^- D_2^- \cdot \ldots \cdot D_N^+ \right]}_{\binom{N}{1} \text{ elements}}$$

$$+ 2 \cdot \underbrace{\left[ D_1^+ D_2^+ \cdot \ldots \cdot D_N^- + D_1^+ D_2^- \cdot \ldots \cdot D_N^- + \ldots + D_1^- D_2^- \cdot \ldots \cdot D_N^+ \right]}_{\binom{N}{2} \text{ elements}} \quad (5)$$

$$+ \ldots$$

$$+ N \cdot \left[ D_1^+ D_2^+ \cdot \ldots \cdot D_N^+ \right]$$

where:

- $N$ – size of population
- $D_i^+$ – probability that the schema $S$ has been matched in observation of $q_i$
- $D_i^-$ – probability that the schema $S$ has not been matched in observation of $q_i$

Consequent lines of the equation (5) correspond to the products of possible numbers of matches and their probabilities. For example, $D_1^- D_2^- \cdot \ldots \cdot D_N^-$ is the probability that no element in the quantum population $Q$ matches the schema. The expression in the second line of the equation (5) is the probability that exactly one (either one) chromosome matched the schema etc. The number of $w$-element subsets of an $N$–element set equals to $\binom{N}{w}$, therefore in consequent lines of the equation $\binom{N}{w}$ elements are added. This allows us to rewrite the equation (5) more concisely, in a manner which is much easier to implement in a programming language[2]:

$$E(L) = \sum_{w=0}^{N} \left( w \sum_{C \in \{ X \in 2^{\{1,\ldots,N\}} : |X| = w \}} \left( \prod_{j \in C} D_j^+ \prod_{k \in \{1,\ldots,N\} \setminus C} D_k^- \right) \right)$$

$$= \sum_{w=0}^{N} \left( w \sum_{C \in \{ X \in 2^{\{1,\ldots,N\}} : |X| = w \}} \left( \prod_{j \in C} M(q_j, S) \prod_{k \in \{1,\ldots,N\} \setminus C} (1 - M(q_k, S)) \right) \right) \quad (6)$$

where:

- $N$ – size of population
- $2^{\{1,\ldots,N\}}$ – set of all subsets of the $\{1, \ldots, N\}$ set
- $\{ X \in 2^{\{1,\ldots,N\}} : |X| = w \}$ – set of all $w$-element subsets of the $\{1,\ldots,N\}$ set

---

[2] Keep in mind that an empty product (product over an empty set) equals 1

The equation (6) presents directly how to compute the expected value $E(L)$ in four loops of programming language, regardless of the population size and the length of chromosomes. The first sum in (6) is taken over all the possible numbers of matches $\{0, ..., N\}$. The second sum is taken over the $w$-element subsets of the $\{1, 2, ..., N\}$ set. Products in the equation (6) correspond to the probability that the given combination matches or does not match, respectively. In the second line of (6), $D_i^+$ and $D_i^-$ have been replaced with $M(q_j, S)$ and $(1 - M(q_k, S))$. Consequently, (6) holds for the number of binary quantum chromosomes matching the schema $S$ in the quantum population $Q = \{q_1, ..., q_N\}$. To calculate the variance $V(L)$ we shall use the following basic properties of a random variable:

$$V(L) = E(L^2) - (E(L))^2$$

$$E(L^2) = \sum_k x_k^2 p(x_k)$$

The expected value of $L^2$ is expressed as:

$$\begin{aligned}
E(L^2) = {} & 0^2 \cdot \left[D_1^- D_2^- \cdots D_N^-\right] \\
& + 1^2 \cdot \left[D_1^+ D_2^- \cdots D_N^- + D_1^- D_2^+ \cdots D_N^- + \ldots + D_1^- D_2^- \cdots D_N^+\right] \\
& + 2^2 \cdot \left[D_1^+ D_2^+ \cdots D_N^- + D_1^+ D_2^- \cdots D_N^- + \ldots + D_1^- D_2^- \cdots D_N^+\right] \\
& + \ldots \\
& + N^2 \cdot \left[D_1^+ D_2^+ \cdots D_N^+\right] \\
= {} & \sum_{w=0}^{N} \left( w^2 \sum_{C \in \{X \in 2^{\{1,\ldots,N\}} : |X|=w\}} \left( \prod_{j \in C} D_j^+ \prod_{k \in \{1,\ldots,N\} \setminus C} D_k^- \right) \right)
\end{aligned} \qquad (7)$$

Consequently, the variance of the number of quantum chromosomes matching the $S$ schema can be calculated with the formula:

$$\begin{aligned}
V(L) = {} & \sum_{w=0}^{N} \left( w^2 \sum_{C \in \{X \in 2^{\{1,\ldots,N\}} : |X|=w\}} \left( \prod_{j \in C} M(q_j, S) \prod_{k \in \{1,\ldots,N\} \setminus C} (1 - M(q_k, S)) \right) \right) \\
& - \left( \sum_{w=0}^{N} \left( w \sum_{C \in \{X \in 2^{\{1,\ldots,N\}} : |X|=w\}} \left( \prod_{j \in C} M(q_j, S) \prod_{k \in \{1,\ldots,N\} \setminus C} (1 - M(q_k, S)) \right) \right) \right)^2
\end{aligned} \qquad (8)$$

## 4. Numerical Experiment

In this section we shall compare building blocks propagation on selected test problem. For our analysis let us consider the following optimization problem. Let $f : [0,200] \mapsto \Re$ denote an objective function, created by a cubic B-splines interpolation for the following knots: {(0, 0), (10, 20), (20, 40), (30, 10), (40, 25), (56, 33), (60, 80), (65, 45), (80, 60), (90, 20), (100, 0), (120, 20), (150, 40), (180, 20), (200, 0)}. The landscape of the objective function and the interpolation knots have been presented in figure 4.

The optimization problem is given as: find $x^* \in [0,200]$, such that $x^* = \arg\max_{x \in [0,200]} f(x)$. The objective function has a global maximum at $x^* = 60.488$, where $f(x^*) = 80.834$. Natural binary coding has been used and the length of chromosomes in both compared algorithms has been set to 20 genes, which results in 0.0001 precision for the considered domain. The binary string which encodes the best possible solution is 01001101011011001110.

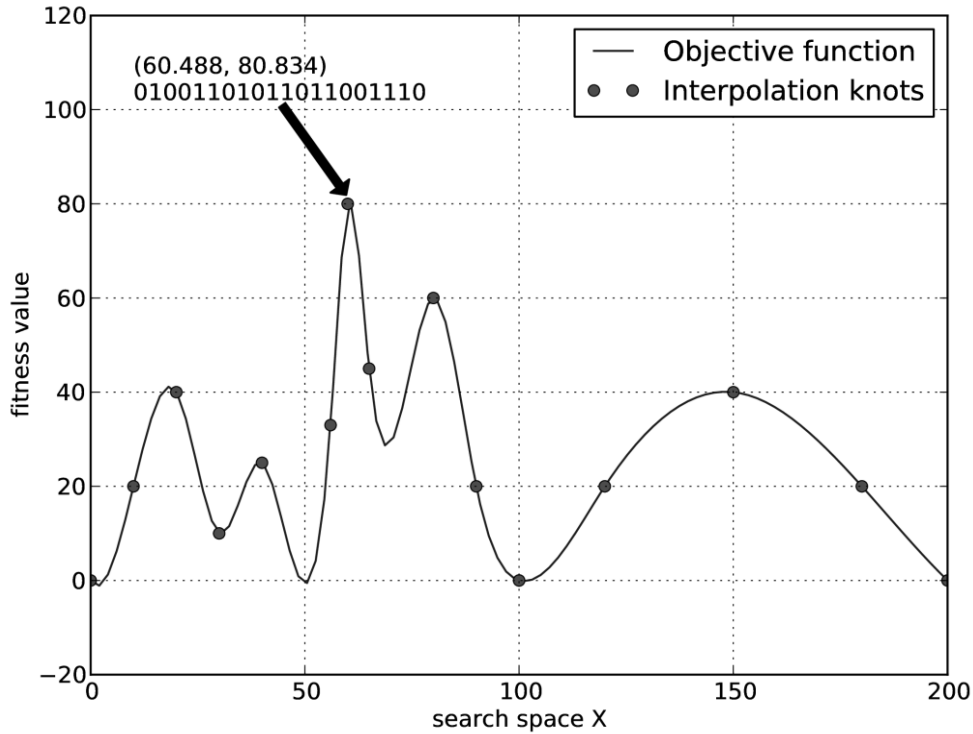

**Fig. 4.** Landscape of the test function

In the experiment two algorithms have been compared for the test problem: Simple Genetic Algorithm and Quantum-Inspired Genetic Algorithm. The algorithms were implemented in the Python programming language. The implementations were in no way optimized to reduce real time of execution, as it is clearly not relevant to the subject of this pa-per. Therefore, the exact information on the execution time was not collected in the experiment. Similarly, no efforts have been taken to perform an accurate computational complexity comparison of the algorithms which is also outside the scope of this study.

The algorithms were started with highly typical parameters given in the literature[4,8,9,14]. The population size was set to 10 individuals. The termination criterion was simply the maximum number of generations: 160. The fitness function corresponds directly to the objective function. Neither scaling nor any other modification of the fitness value has been applied. In Simple Genetic Algorithm one-point crossover has been used. The probability of crossover was $P_c$=0.9. Uniform mutation has been applied and the probability of mutation was $P_m$=0.005, which results in mutation of one gene in the population per one generation on average. Fitness-proportionate (roulette wheel) selection method has been applied.

In Quantum-Inspired Genetic Algorithm parameters given originally in [8] have been used. The crucial parameters of this algorithm are rotation angles given in so-called lookup table. The rotation angles were also taken directly from [8]. Despite the rotation angles were initially adapted for the other type of problem (i.e. combinatorial optimization) the algorithm performs also very well on the numerical optimization. The lookup table is the only special parameter for the QIGA algorithm, apart from the number of chromosomes and their length.

Assuming that the global optimum of our test functions is known, we can select a building block (a short, low order, high fitness schema) and test its propagation in consecutive generations. Such methodology has been already used in the past years for analysis of several evolutionary algorithms[12]. In our experiment, schemata of maximum defining length 4 and maximum order 5 have been considered as possible building blocks. Selected best schemata and their fitness are given in table 2. Average fitness of all $2^{20}$ chromosomes is 25.068. For further analysis the schema 01001*************** has been selected as a building block.

**Table** 2

Selected best schemata and their fitness for the test problem

| schema | fitness value |
|---|---|
| 01001*************** | 67.878 |
| 01*01*************** | 57.797 |
| *10011************** | 56.992 |
| 010*1*************** | 51.361 |
| *10*11************** | 44.988 |
| *100*1************** | 43.133 |

## 5. Results

In figure 5, efficiency comparison of the algorithms has been presented. The fitness of the best (maximum) individuals in populations of consecutive generations has been considered. The plot has been created for an average over 10 runs of the algorithms. In 50 generations the Quantum-Inspired Genetic Algorithms always found the optimum solution. It is straightforward to notice that Quantum-Inspired Genetic Algorithm outperforms the classical Simple Genetic Algorithm for this test problem easily. For this very simple test problem, the poor results of Simple Genetic Algorithm in this case arise from relatively small size of population (10 individuals), no elitism, fitness-proportionate selection method and no scaling of the fitness value.

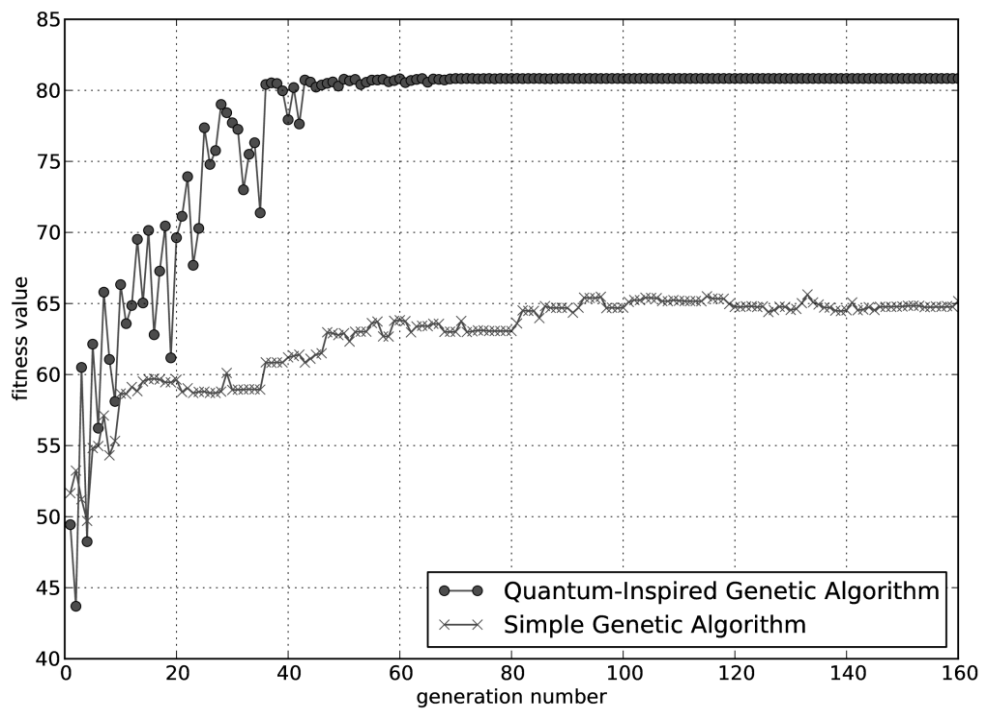

**Fig. 5.** Efficiency comparison of two algorithms for the test problem

Figure 6 presents the building block propagation comparison for the selected building block 01001***************. The plot has been created for an average over 10 runs of the algorithms. In 80 generations of the Simple Genetic Algorithm approximately four chromosomes containing the building block have evolved. This is subject to the Holland's schema theorem [9], which gives the lower bound of the propagation in Simple Genetic Algorithm. The number of approximately four chromosomes containing the building block remained stable in subsequent generations, fluctuating slightly.

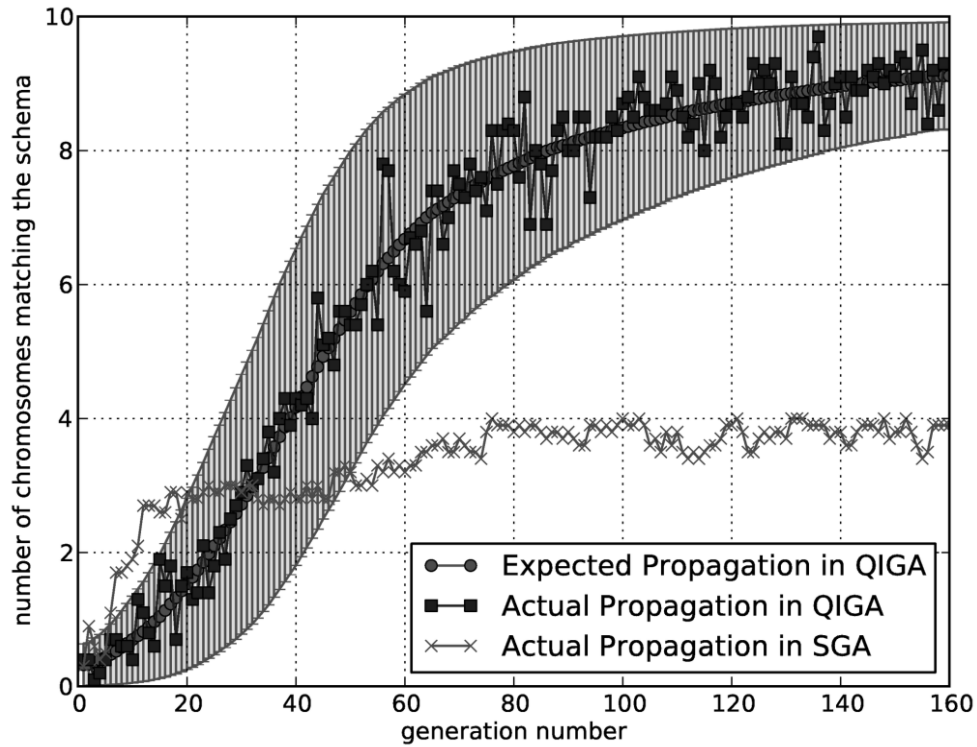

**Fig. 6.** Building block propagation comparison in QIGA and SGA

Results for Quantum-Inspired Genetic algorithm has been depicted in figure 6 as a region representing expected number and variance of chromosomes containing the considered building block. The expected number and the variance have been calculated with equations (6) and (8). The actual number of the building blocks observed in consequent generations of populations in Quantum-Inspired Genetic Algorithm has been also presented in figure 6. The actual propagation verifies that the formulas (6) and (8) have been devised correctly. In 80 generations of the Quantum-Inspired Genetic Algorithm approximately 8 chromosomes matching the schema have evolved. The variance of this value is about 2 in that generation. Therefore, it outperforms the Simple Genetic Algorithm even in a bad case. In subsequent generations the expected number levelled off and it reached approximately 9 chromosomes in 160 generation. At the same time the variance has been decreasing progressively as the quantum population converged to the optimum solution and the uncertainty in quantum chromosomes has been reduced. Moreover, it is easy to notice in figure 6 a monotonous regularity of building block propagation in Quantum-Inspired Genetic Algorithms compared to Simple Genetic Algorithm.

## 6. Discussion

To illustrate the behaviour of quantum genetic algorithm in consequents generations for the test problem, let us consider the quantum chromosome $q = g_1g_2g_3g_4g_5$ of five binary quantum genes. The state of such quantum chromosome can be simply depicted by five arrows, this notation has been already introduced in section 2 of this paper. Figure 7 presents an illustration of six different states of the $q$ chromosome.

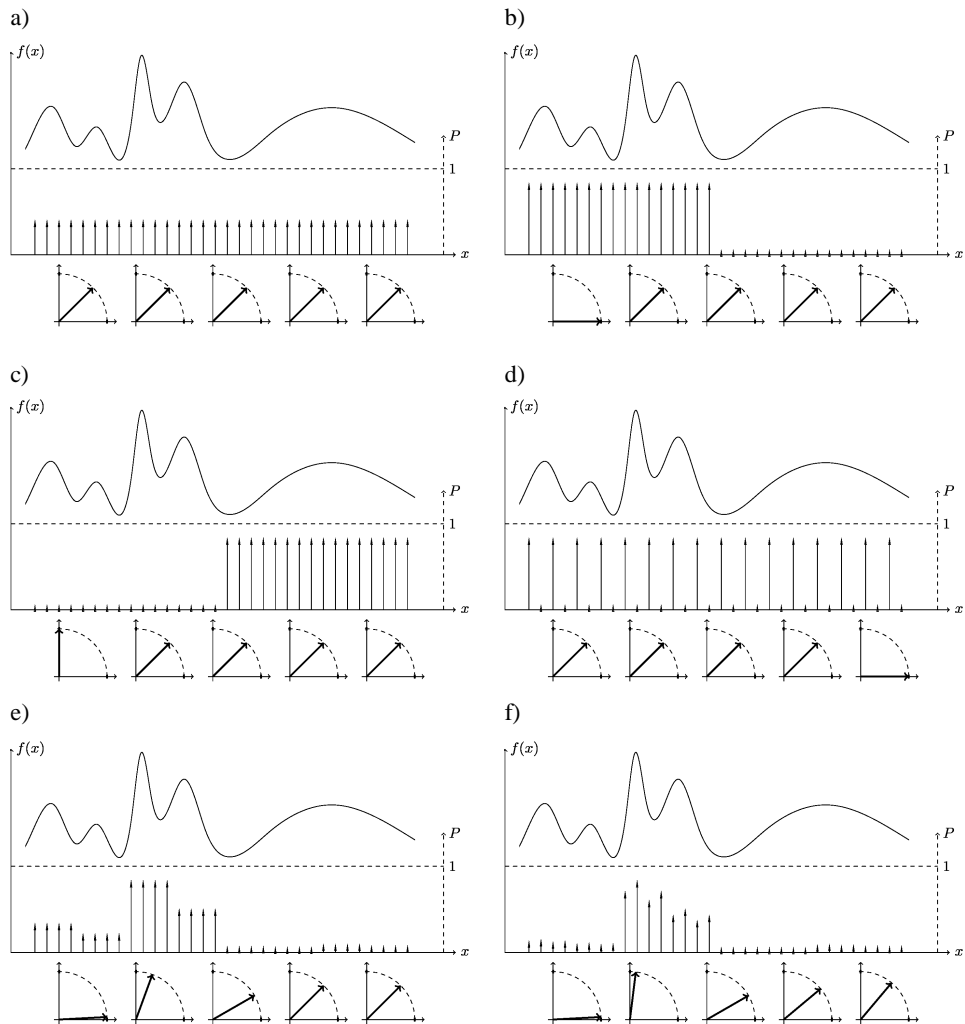

**Fig. 7.** Illustration of selected binary quantum chromosomes of 5 genes (diagonal arrows below the plots) and corresponding probability distributions of probing the search space (comb plots).

In the beginning of the algorithm the quantum population is initialized by quantum chromosomes which "cover" the whole search space with equal probability. In this situation the algorithm performs a global search in the whole domain and it has been depicted in figure 7a, which presents the initial state of the quantum chromosome: $\left(\frac{\sqrt{2}}{2}|0\rangle + \frac{\sqrt{2}}{2}|1\rangle\right)\left(\frac{\sqrt{2}}{2}|0\rangle + \frac{\sqrt{2}}{2}|1\rangle\right)\left(\frac{\sqrt{2}}{2}|0\rangle + \frac{\sqrt{2}}{2}|1\rangle\right)\left(\frac{\sqrt{2}}{2}|0\rangle + \frac{\sqrt{2}}{2}|1\rangle\right)\left(\frac{\sqrt{2}}{2}|0\rangle + \frac{\sqrt{2}}{2}|1\rangle\right)$. Such state of the quantum chromosome is similar to the schema ***** (wildcards on all positions). Figures 7b and 7c illustrate probability distributions encoded by quantum chromosomes with no uncertainty involved in the first quantum gene. Figure 7b corresponds to quantum chromosome $|0\rangle \left(\frac{\sqrt{2}}{2}|0\rangle + \frac{\sqrt{2}}{2}|1\rangle\right)\left(\frac{\sqrt{2}}{2}|0\rangle + \frac{\sqrt{2}}{2}|1\rangle\right)\left(\frac{\sqrt{2}}{2}|0\rangle + \frac{\sqrt{2}}{2}|1\rangle\right)\left(\frac{\sqrt{2}}{2}|0\rangle + \frac{\sqrt{2}}{2}|1\rangle\right)$. The probability distribution of this chromosome covers the first half of the search space and it corresponds to the schema 0****. Consequently, the algorithm samples only this region for such individual. On the contrary, the quantum chromosome $|1\rangle \left(\frac{\sqrt{2}}{2}|0\rangle + \frac{\sqrt{2}}{2}|1\rangle\right)\left(\frac{\sqrt{2}}{2}|0\rangle + \frac{\sqrt{2}}{2}|1\rangle\right)\left(\frac{\sqrt{2}}{2}|0\rangle + \frac{\sqrt{2}}{2}|1\rangle\right)\left(\frac{\sqrt{2}}{2}|0\rangle + \frac{\sqrt{2}}{2}|1\rangle\right)$ corresponds to the schema 1**** and to the probability distribution that covers only the second half of the search space. In this situation (figure 7c) only the second half of the search space is sampled by the algorithm. In figure 7d the binary quantum chromosome $\left(\frac{\sqrt{2}}{2}|0\rangle + \frac{\sqrt{2}}{2}|1\rangle\right)\left(\frac{\sqrt{2}}{2}|0\rangle + \frac{\sqrt{2}}{2}|1\rangle\right)\left(\frac{\sqrt{2}}{2}|0\rangle + \frac{\sqrt{2}}{2}|1\rangle\right)\left(\frac{\sqrt{2}}{2}|0\rangle + \frac{\sqrt{2}}{2}|1\rangle\right)|0\rangle$ has been depicted (no uncertainty in the last gene). Such quantum chromosome corresponds to the schema ****0. Assuming that all the elements of our search space can be enumerated and therefore they create a sequence, such probability distribution samples only the odd elements of the domain. Correspondingly, the binary quantum chromosome $\left(\frac{\sqrt{2}}{2}|0\rangle + \frac{\sqrt{2}}{2}|1\rangle\right)\left(\frac{\sqrt{2}}{2}|0\rangle + \frac{\sqrt{2}}{2}|1\rangle\right)\left(\frac{\sqrt{2}}{2}|0\rangle + \frac{\sqrt{2}}{2}|1\rangle\right)\left(\frac{\sqrt{2}}{2}|0\rangle + \frac{\sqrt{2}}{2}|1\rangle\right)|1\rangle$ probes only the even elements of the domain. In consequent iterations of the algorithm the probability distribution encoded in binary quantum chromosomes are adjusted to match the landscape of the fitness function and eventually only regions of the highest fitness values are sampled by the algorithm (figures 7e and 7f).

## 7. Conclusions

In this paper a method of building blocks propagation analysis in quantum genetic algorithms has been presented. A random variable corresponding to the number of quantum chromosomes matching a schema has been introduced. Basic measures of the random variable have been calculated. The proposed method of analysis based on the introduced random variable and its measures is the main contribution of this paper. The analysis has been illustrated on the selected one-dimensional test function optimization problem. Building block propagation in Simple Genetic Algorithm and Quantum-Inspired Genetic Algorithm has been compared for selected test function in the numerical experiment. Quantum-Inspired Genetic Algorithm outperformed Simple Genetic Algorithm for the considered test problem. Much better propagation of building block has been observed in Quantum-Inspired Genetic Algorithm.

Further research in the area of theoretical studies of the quantum genetic algorithms includes a possible broad generalization of Holland's schema theorem to the new class of algorithms. First of all, the analysis performed in this paper should be extended for other test problems, other building blocks and other quantum genetic algorithms (e.g. GAQPR, NQEA). Moreover, it should be investigated how selected parameters of the algorithms influence the building blocks propagation and its uncertainty.

Recenzent: